\documentclass[conference]{IEEEtran}
\IEEEoverridecommandlockouts
\usepackage{cite}
\usepackage{amsmath,amssymb,amsfonts}
\usepackage{algorithmic}
\usepackage{graphicx}
\usepackage{textcomp}
\usepackage{xcolor}
\usepackage{url}
\usepackage{balance}
\def\BibTeX{{\rm B\kern-.05em{\sc i\kern-.025em b}\kern-.08em
    T\kern-.1667em\lower.7ex\hbox{E}\kern-.125emX}}
\usepackage[spanish]{babel}

\begin{document}

\title{Reducción de ruido por medio de \textit{autoencoders}: caso de estudio con la señal GW150914\\}

\author{
\IEEEauthorblockN{1\textsuperscript{nd} Fernanda Javiera Zapata Bascuñán}
\IEEEauthorblockA{\textit{Universidad Nacional del Comahue} \\
Neuquén, Argentina \\
fernanda.zapata@fain.uncoma.edu.ar}

\and

\IEEEauthorblockN{2\textsuperscript{st} Darío Fernando Mendieta}
\IEEEauthorblockA{\textit{Universidad Nacional del Comahue} \\
Neuquén, Argentina \\
dario.mendieta@fain.uncoma.edu.ar}

}

\maketitle

\begin{abstract}

Este breve estudio se enfoca en la aplicación de \textit{autoencoders} para mejorar la calidad de señales de pequeña amplitud, como el caso de eventos gravitacionales. Se entrenó un \textit{autoencoder} preexistente utilizando datos de eventos cósmicos, optimizando su arquitectura y parámetros. Los resultados muestran un aumento significativo de la relación señal a ruido en las señales procesadas, demostrando el potencial de los \textit{autoencoders} en el análisis de pequeñas señales con múltiples fuentes de interferencia.
\end{abstract}


\section{Introducción}
Con el avance de la detección y el análisis de ondas gravitacionales, el estudio y la aplicación de técnicas de procesamiento de señales se han vuelto esenciales para mejorar la detección de eventos y la calidad de los datos observados. En este contexto, los \textit{autoencoders} han surgido como una herramienta prometedora [1] para la reducción de ruido en señales, permitiendo la obtención de información relevante en entornos bajo múltiples fuentes de interferencia y ruido.

Este trabajo se enfoca en la aplicación de \textit{autoencoders} como una metodología efectiva para reducir el ruido presente en señales gravitacionales, con un énfasis especial en un caso de estudio: la señal GW150914. Esta señal, resultado de la fusión de agujeros negros binarios, representa un hito histórico en la detección de ondas gravitacionales y proporciona una base sólida para investigar y validar la eficacia de los \textit{autoencoders} en un contexto real. Además, a diferencia de otros eventos publicados posteriormente, este evento se destaca por tener la relación señal/ruido más fuerte [2] [3], siendo un buen candidato para el propósito de éste estudio.

A través de este estudio, se busca demostrar cómo la implementación de un \textit{autoencoders} puede resultar efectivo en comparación con las distintas técnicas aplicadas en los tutoriales de LIGO Open Science Center [4] o la depuración de la señal como se demuestra en [2]. Cabe mencionar que, para abordar otros eventos, sería necesario generar un \textit{dataset} específico.

\section{Generalidades}

\subsection{Codificadores Automáticos}
Un \textit{autoencoder} es un tipo de red neuronal de comprensión donde las funciones de compresión y decompresión tienen ciertas características: son específicas para a los datos de entrada, cada una tiene relacionada con una función de pérdida y aprenden de forma automática y no supervisada, ya que los datos no requieren estar etiquetados.
El proceso de autocodificación involucra dos etapas principales: el codificador y el decodificador. El codificador comprime los datos desde un espacio de mayor dimensión a un espacio de menor dimensión (también llamado espacio latente), mientras que el decodificador realiza la operación opuesta, es decir, convierte el espacio latente de nuevo al espacio de mayor dimensión. La función del decodificador es asegurar que el espacio latente capture la mayor cantidad de información posible del espacio de datos, de esta forma la salida resulta en una representación cercana a la entrada original que se proporcionó al autoencoder.
\begin{figure}[htbp]
\centerline{\includegraphics[scale=0.6]{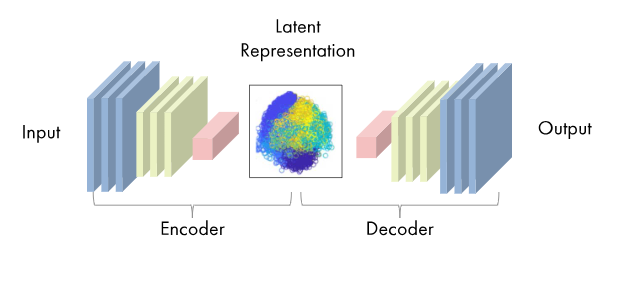}}
\caption{Estructura esquemática de un \textit{autoencoder}. $^1$}
\label{fig 1}
\end{figure}
\footnote{Imagen recuperada de https://la.mathworks.com/discovery/autoencoder.html}
\\El entrenamiento de un autocodificador se puede pensar como la optimización de la función de costo asociada a las capas neuronales escogidas.

Existen múltiples formas de implementación y arquitectura interna de los \textit{autoencoder} tales como \textit{Dense Neural Networks (DNNs)}, \textit{Recurrent Neural Networks (RNNs) LSTMs or GRUs,} Capas Convolucionales y \textit{Spiking Neural Networks (SNN)}. Se optó por trabajar con la primera de ellas ya que los datos  con frecuencia no presentan una dependencia temporal explícita y no se observa una estructura de secuencia en los patrones subyacentes. Por lo tanto, para el caso de los autocodificadores, como en el presente enfoque, las \textit{DNNs} resultan ser una elección adecuada y ventajosa.

\subsection{Caso de estudio: Señal GW150914}
El caso de estudio se centra en el evento cósmico observado el 14 de septiembre del año 2015 a las 09:50:45 UTC. Según [5] el procedimiento para detectar la señal comenzó con búsquedas de baja latencia de transitorios gravitacionales genéricos en los interferómetros de Hanford y Livingston. Una vez detectado el evento se utilizó una técnica de coincidencia de filtros con plantillas  relativistas analíticas obtenidas de librerías como PyCBC y GstLAL y de las cuales se obtuvo a la señal GW150914 como la más probable en el evento. También se obtuvieron otros parámetros de la señal como la relación señal a ruido, siendo la misma de 24, además de la identificación de ruido no Gaussiano en la señal [5]. Ésto último de gran interés para el presente trabajo, por la relación de dependencia que puede existir entre los datos y el tipo de procesamiento de la señal.\\

\begin{figure}[htbp]
\centerline{\includegraphics[scale=0.37]{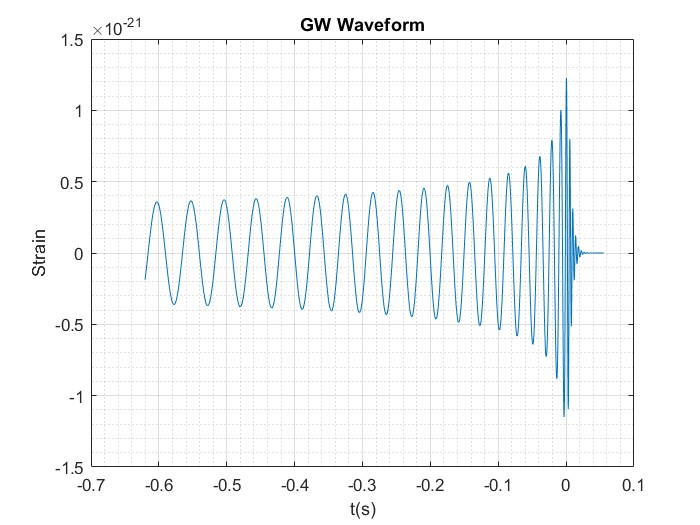}}
\caption{Plantilla GW150914 proporcionada en \textit{gwosc}.}
\label{fig2}
\end{figure}

El evento involucró la fusión de dos agujeros negros. El agujero negro primario tenía una masa estimada entre 32 a 41 masas solares, momento angular menor a 0.7 y el agujero negro secundario con masa entre 25 a 33 masas solares, momento angular menor a 0.9. La fusión de ambos tuvo una duración de 200ms y el agujero negro resultante tiene una masa estimada de entre 60 a 70 masas solares. Del evento se observó en los detectores un pico de deformación de 1e-21 unidades relativas. La señal de deformación se observa en la Fig. 2.\\

La deformación (\textit{strain}) es una medida del cambio en la distancia entre dos puntos en el espacio debido a la influencia de la onda gravitacional y es la magnitud detectada por los interferómetros de ondas gravitacionales. Esta deformación toma forma de una señal \textit{chirp}, cuya frecuencia y amplitud contiene información del sistema binario. La forma de onda depende de la masa del sistema binario según [5],
\begin{equation} M = (m_1m_2)^{3/5}/(m_1 + m_2)^{1/5},\end{equation} 
la relación de masas simétrica,
\begin{equation}\eta = (m_1m_2)/(m_1 + m_2)^2 \end{equation}
y el momento angular de los objetos compactos:
\begin{equation}\chi_{1,2} = cS_{1,2}/Gm_{21,2}\end{equation}
Estas relaciones fueron consideradas al momento de generar el \textit{dataset} ya que resulta esencial incorporar modelos analíticos que abarquen la simulación de los parámetros anteriores. 

 \subsection{Señales de los detectores}
Las señales reales del evento obtenidas de los detectores se encuentran disponibles en la base de datos de LIGO y abierto al público en general en [4]. Las señales se encuentran en formato hdf5 y contemplan 32 y 4096 segundos alrededor del evento. Estas señales han sido muestreadas a 16384Hz y 4096Hz respectivamente. En la Fig. 3. se muestra, para cada detector, 10 segundos alrededor del evento.

\begin{figure}[htbp]
\centerline{\includegraphics[scale=0.35]{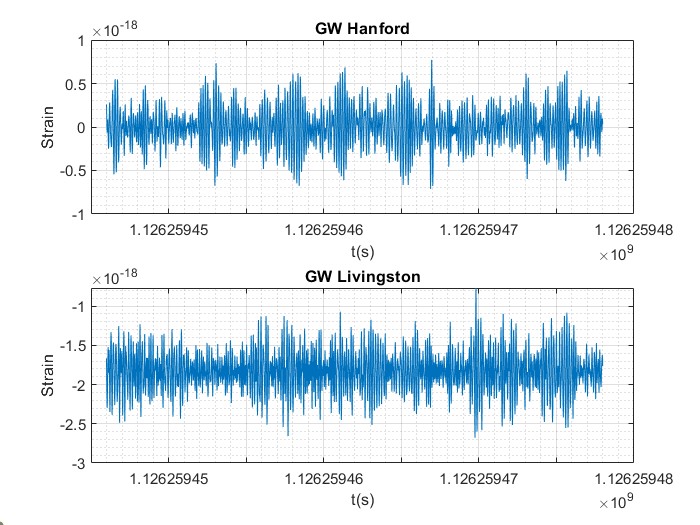}}
\caption{GW150914 obtenidas en los detectores}
\label{fig3}
\end{figure}

\section{Creación del banco de plantillas}

Para entrenar el \textit{autoencoder}, se procedió a crear un banco de plantillas utilizando modelos fenomenológicos de ondas gravitacionales por medio de las librerías \textit{PyCBC} y \textit{LALsuite}. Los modelos empleados resultan ser cruciales en la descripción de las señales emitidas por sistemas astrofísicos, específicamente, agujeros negros binarios (BBH) [6], en procesos de colisión y fusión. Entre los modelos empleados en el \textit{dataset} se destacan \textit{IMRPhenom, SEOBNRv4} y \textit{Taylorv4} [7,8], los cuales ofrecen representaciones teóricas de las formas de onda que emergen de dichos eventos astrofísicos y a diferencia de otros consideran características similares a GW150914 como el momento angular de los agujeros negros.

\begin{figure}[htbp]
\centerline{\includegraphics[scale=0.4]{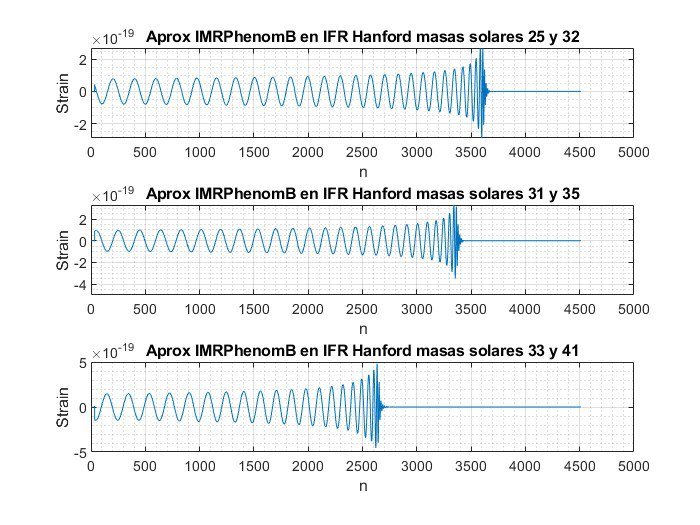}}
\caption{Tres señales generadas a partir del modelo IMRPhenom con masas variables dentro de los rangos admitidos.}
\label{fig4}
\end{figure}

Para formar el conjunto de señales, se consideraron las propiedades inherentes del evento, específicamente en relación a la deformación. Sólo las masas fueron alteradas para crear el conjunto de datos, mientras que otras características como el momento angular, la inclinación y la frecuencia mínima se mantuvieron sin cambios. No obstante, se ajustaron las masas de los agujeros negros de acuerdo a los parámetros previamente identificados, con el propósito de generar diversas plantillas que abarcaran la descripción completa del evento. 

Para ello, se escogió un valor de spin máximo de cada agujero negro, 0.7 y 0.9 respectivamente, la aparición en el cielo según [10], y la variación de masas de los agujeros negros con paso 0.5 masas solares entre los mínimos y máximos previamente descritos. La Fig 5. muesra una tabla que representa las señales obtenidas para el conjunto de plantillas.
\begin{figure}[htbp]
\centerline{\includegraphics[scale=0.4]{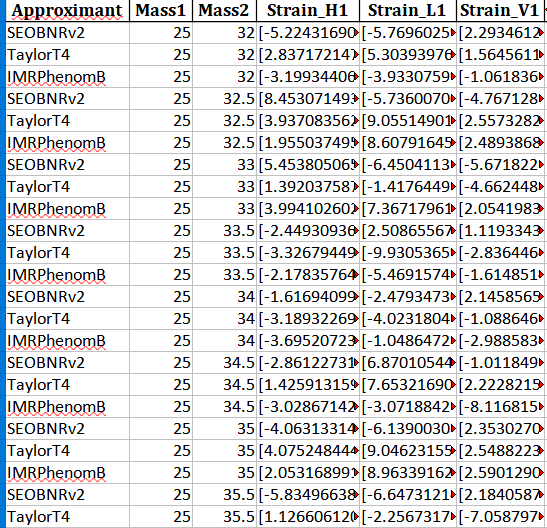}}
\caption{Banco de plantillas.}
\label{fig5}
\end{figure}

Después de la creación del conjunto de señales, se procedió a simular la señal de onda recepcionada por cada detector en particular. Los interferómetros de Hanford, Livingston y Virgo presentan diferencias en sus características y ubicaciones geográficas. Debido a estas variaciones, la deformación registrada por cada interferómetro puede diferir para un mismo evento. En el contexto de este estudio, se consideró la respuesta de cada señal analítica a los detectores para generar un conjunto de señales que reflejara las distintas observaciones posibles. 

De esta manera, se obtuvieron un total de 2261 señales de longitud 4511. Se sometió al conjunto de señales a un proceso de normalización que tenía como objetivo estandarizar la longitud de todas las señales para asegurar el buen desempeño del algoritmo posterior. Para lograr esto, se determinó la longitud de la señal más larga dentro del \textit{dataset} y se ajustaron las demás señales para que todas tuvieran la misma longitud. Esto se logró agregando ceros al final de las señales, sin modificar el momento de inicio de cada una. 
Esta evaluación se considera esencial para determinar la capacidad del \textit{autoencoder} para identificar la señal en los datos recopilados por los interferómetros.

En la Fig. 3 se observa una muestra del \textit{dataset} que contempla un método analítico, uno de los tres detectores mencionados y la modificación de las masas de los agujeros negros involucrados. 

Mediante estas consideraciones, se logró generar un \textit{dataset} acorde a las características deseadas, y a la vez se introdujo la variabilidad de las masas y de los detectores: Hanford (H1), Livingston (L1) y Virgo (V1). 

Una vez definidos los parámetros, la generación del dataset demoró aproximadamente 3 horas. Los archivos se exportaron en formato '.txt' con el fin de preservar la mayor cantidad de decimales de precisión. 

\section{Implementación del Autocodificador}

 \subsection{Sparse Autoencoder}
 \begin{figure}[htbp]
\centerline{\includegraphics[scale=0.35]{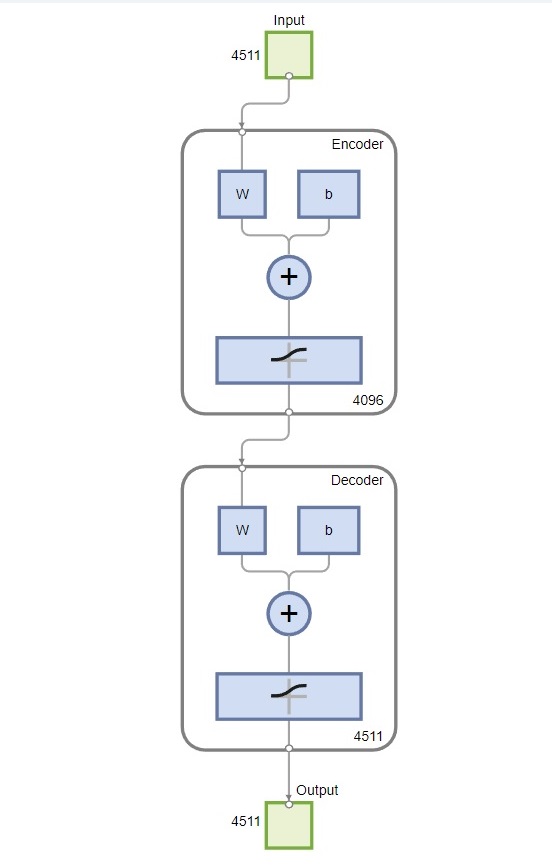}}
\caption{Arquitectura del autoencoder}
\label{fig6}
\end{figure}

Un autocodificador disperso o \textit{sparse autoencoder} es un \textit{autoencoder} cuyo criterio de entrenamiento implica una penalización por dispersión $\Omega(h)$ en la capa de código \textit{h}, además del error de reconstrucción:
\begin{equation}
L(x, g(f(x))) + \Omega(h)
\end{equation} 

donde $g(h)$ es la salida del decodificador y típicamente tenemos $h = f(x)$, la salida del codificador.
Los \textit{autoencoders} con esparcidad se utilizan comúnmente para aprender características para tareas de clasificación. Sin embargo, un \textit{autoencoder} que ha sido sometido a regularización para obtener dispersión es capaz de reconocer y adaptarse a las características estadísticas distintivas del conjunto de datos en el que fue entrenado. Este enfoque difiere de simplemente comportarse como una función identidad. En lugar de ello, busca que el modelo ajuste sus parámetros de manera que refleje las propiedades fundamentales del conjunto de datos.

La penalización $\Omega(h)$ se puede entender simplemente como un término de regularización añadido a una red neuronal \textit{feedforward} cuya tarea principal es copiar la entrada en la salida y posiblemente también realizar alguna tarea supervisada que dependa de estas características dispersas. A diferencia de otros regularizadores, como la disminución de peso (\textit{weight decay}) $\beta$, no tiene una interpretación bayesiana directa, mas bien se basa según [11,12,13] en una interpretacion MAP a la inferencia bayesiana, donde la penalización de regularización adicional corresponde a una distribución de probabilidad a priori sobre los parámetros del modelo.

\begin{equation}
\varrho_i = \frac{1}{n} \sum_{j=1}^{n} z_j^{(1)}(x_j)
\end{equation} 

Otro parámetro presente en el \textit{sparse autoencoder} es la regularización L2 (Eq 6). También conocido como \textit{Ridge Regression}, éste modelo de regresión, añade la magnitud al cuadrado de los coeficientes de cada capa como término de penalización a la función de pérdida. Se utiliza comúnmente como un mecanismo que orienta al modelo a aprender representaciones más cercanas a la distribución deseada, en éste caso, del evento gravitacional en estudio.

\begin{equation}
\Omega_{weights} = \frac{1}{2} \sum_{l=1}^{L}\sum_{j=1}^{n_l}\sum_{i=1}^{k_l}(w_{ji}^{(l)})^2
\end{equation} 

La medida de divergencia de esparcidad se implementa mediante la divergencia de Kullback-Leibler (Eq. 7). 
En el contexto del aprendizaje automático y la inferencia estadística, la divergencia de Kullback-Leibler se utiliza frecuentemente para evaluar la disparidad entre dos distribuciones de probabilidad. Esta diferencia puede manifestarse como la discrepancia entre la distribución real de los datos y la distribución que el modelo intenta capturar, la cual se desea minimizar (Eq. 4). De este modo, se busca que el \textit{autoencoder} examine las características intrínsecas de los datos analizados, logrando capturar la distribución de probabilidad del modelo de manera precisa. Es por eso que suele utilizarse en entornos de aprendizaje no supervisado, como el agrupamiento, la reducción de dimensionalidad y la estimación de densidad.

\begin{equation}
\Omega(h)_{sparsity} = \sum_{i=1}^{n} KL(\rho || \varrho)
\end{equation}

\subsection{\textit{Fine-Tunning}}

Para un acercamiento a los \textit{sparce autoencoders}, se decidió implementar el autoencoder preexistente de MATLAB. En otros trabajos como en [13] se observa que para \textit{denoising autoencoders} del tipo convolucional, menor cantidad de capas suele se una buena aproximación a la depuración de señales. También en trabajos que involucran la depuración de ruido en imagenes como en [15] se ha podido observar que la eficacia aumenta con la disminución de parámetros entrenables. Siguiendo en éste presedente, se optó por seleccionar la arquitectura ya existente y llevar a cabo un ajuste fino con el objetivo de evaluar la eficacia del algoritmo. \\

Se consideraron las siguientes características para el entrenamiento:

\begin{itemize}
  \item Unidades ocultas: 4096
  \item \textit{Max epoch}: 100
  \item \textit{L2WeightRegularization}: 0,001
  \item \textit{SparsityRegularization}: 4
  \item \textit{SparsityProportion}: 0,05
  \item \textit{EncoderTransferFunction}: Sigmoide logarítmica.\\
\end{itemize}

\begin{figure}[htbp]
\centerline{\includegraphics[scale=0.4]{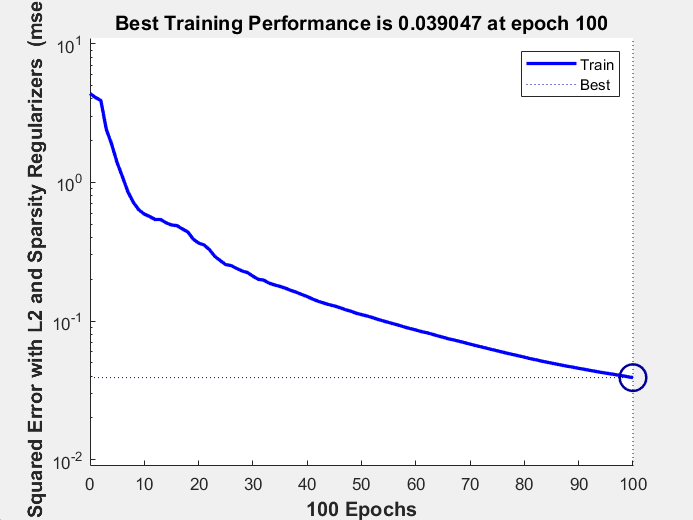}}
\caption{Curva de entrenamiento.}
\label{fig7}
\end{figure}
Se realizaron algunas evaluaciones empíricas con respecto al tamaño del \textit{dataset}, la cantidad de capas ocultas, la cantidad de épocas y la esparcidad. Estas evaluaciones se realizaron con el propósito de asegurar un rendimiento fluido y eficiente del algoritmo en el dispositivo de entrenamiento utilizado. Además, se tuvo especial cuidado en no superar la longitud de los datos de entrada, porque como observó en trabajos anteriores [8], si las capas ocultas superan la longitud de los datos de entrada, entonces se corre el riesgo de generar un sistema que sólo repita la entrada o sobreajuste a los datos de entrenamiento.
Es importante destacar que este cuidado no se considera simplemente como una precaución superficial, ya que se reconoce que \textit{autoencoders} regularizadores, como el utilizado en este estudio, pueden, incluso siendo sobrecompletos, aprender información valiosa sobre la distribución de datos, incluso si la capacidad del modelo es lo suficientemente grande como para aprender una función de identidad trivial.

 \subsection{Entrenamiento}
El proceso de entrenamiento se realizó luego de ajustar los parámetros de las unidades ocultas, la cantidad de épocas, la función de costo empleada y las capas de normalización. 
En cuanto a la implementación, la Fig. 5 muestra la arquitectura del \textit{autoencoder} utilizada en el estudio.
La estructura interna del autoencoder cuenta con tres capas: una capa de entrada, una de salida y una de unidades ocultas.

En términos de \textit{hardware}, el algoritmo se ejecutó en una GPU GTX1060 con 32GB de RAM, funcionando en conjunto con un procesador Intel I5.

Se realizaron dos entrenamientos a fin de encontrar empíricamente la configuración de capas de activación que resultara más acorde al caso de estudio.
\begin{figure}[htbp]
\centerline{\includegraphics[scale=0.38]{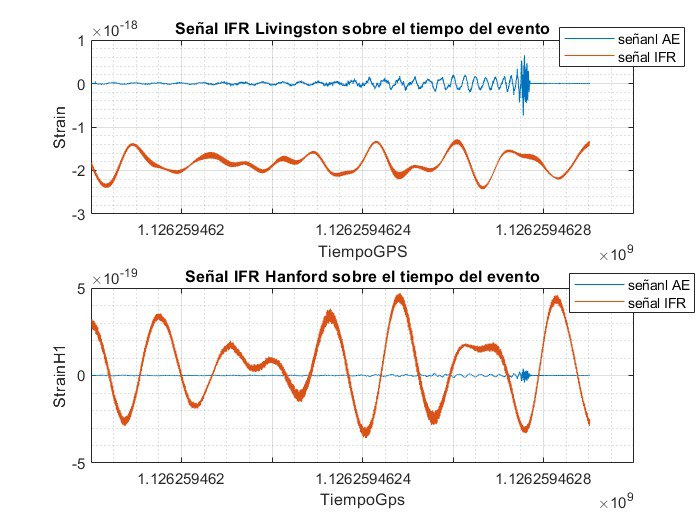}}
\caption{Señales procesadas con la configuración autoencoder I.}
\label{fig9}
\end{figure}

El primer entrenamiento contempló la función de activación \textit{purelin}. La función \textit{pure line} es una función de activación que mapea los valores de entrada sobre una función lineal.
\\De éste primer entrenamiento se obtuvo el \textit{autoencoder I}.

En el segundo entrenamiento se consideraron los parámetros de entrenamiento anteriores y se modificó únicamente la función de activación del decodificador. Se escogió para este entrenamiento la función \textit{logsig} con el fin de preservar la simetría en la arquitectura entrada-salida. La función  sigmoide logarítmica es una función de activación que mapea los valores de entrada entre valores de [0,1]. 
\\De éste segundo entrenamiento se obtuvo el \textit{autoencoder II}.

\begin{figure}[htbp]
\centerline{\includegraphics[scale=0.38]{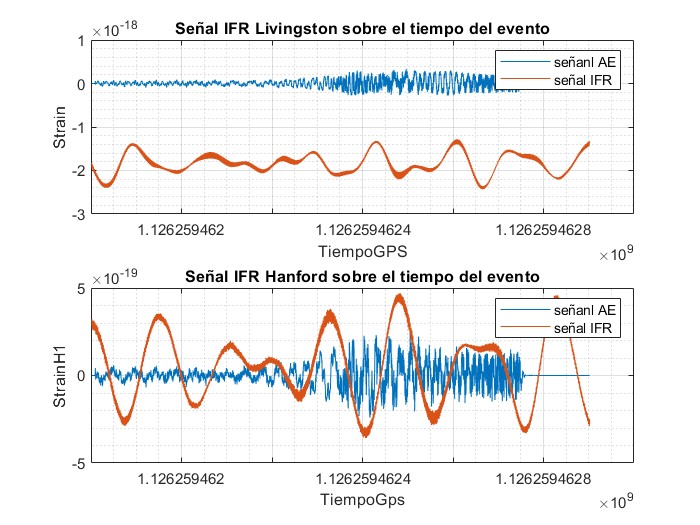}}
\caption{Señales procesadas con la configuración \textit{autoencoder II}.}
\label{fig8}
\end{figure}

\section{Resultados}
Para analizar el desempeño del \textit{autoencoder} se contempló el uso de las señales reales obtenidas de la detección del evento en los interferómetros de Hanford y Livingston. Ambas señales fueron procesadas por los \textit{autoencoders} mencionados en el apartado anterior con el propósito de comparar las señales y seleccionar el autoencoder que mejor lograra eliminar el ruido asociado al evento.

Las Fig.8 y Fig.9 muestran las señales obtenidas de los detectores (naranja) y las mismas señales procesadas por el \textit{autoencoder} (azul). La Fig. 8 muestra los resultados del \textit{autoencoder I}, mientras que la Fig. 9 muestra los resultados del \textit{autoencoder II}.

Al analizar las figuras, se puede notar las siguientes observaciones. En primer lugar, se aprecia la eliminación de la componente de continua en la señal de Livingston en ambas arquitecturas propuestas. Además, en ambas figuras, se evidencia una coincidencia temporal en el inicio del evento para ambas configuraciones, lo que lleva a inferir que efectivamente el \textit{autoencoder} está detectando la señal del evento.

También se observa para el \textit{autoencoder II} que la señal de salida no alcanza una representación acorde a la señal buscada. Una posible causa del bajo rendimiento de ésta configuración puede ser la saturación de gradientes, lo cual suele suceder si los datos tienen valores cercanos a 0 o 1 y resulta común en configuraciones que contemplen sigmoides logarítmicas como funciones de activación.

Es por esto que se concluye que de las dos arquitecturas propuestas para el \textit{autoencoder}, la primera arquitectura (\textit{autoencoder I}) resultó ser la más eficiente en la reducción del ruido presente en la señal detectada. 

En la Figura 10 se presenta la densidad espectral (ASD) para las señal detectadas por Hanford y Livingston procesadas mediante el \textit{autoencoder I}. Al comparar la ASD de las señales originales provenientes de los detectores con la ASD de la señal después de la aplicación del \textit{autoencoder}, se observa una marcada reducción en la amplitud de las ASD de las señales detectadas en determinadas frecuencias significativas por encima de los 1000 Hz, así como una disminución general en las frecuencias bajas con respecto a la ASD de las señales obtenidas tras el \textit{autoencoder I}.
\balance
La relación señal a ruido (SNR) se determinó mediante el uso de la señal detectada por el interferómetro y la señal filtrada por el \textit{autoencoder I}. Los resultados revelaron una mejora notable en la SNR, con valores de -2.55 dB para la señal de Hanford y -24.16 dB para la señal de Livingston. Éstos datos no solo subrayan la capacidad del \textit{autoencoder I} para significativamente mejorar la relación señal a ruido, sino que también reflejan la coherencia con la relación de señal a ruido originalmente calculada en [5]. Este hallazgo refuerza la idea de eficacia del \textit{autoencoder I} en la mejora de la calidad de la señal, respaldando así la validez de los resultados obtenidos y su coherencia con análisis anteriores.

\begin{figure}[htbp]
\centerline{\includegraphics[scale=0.4]{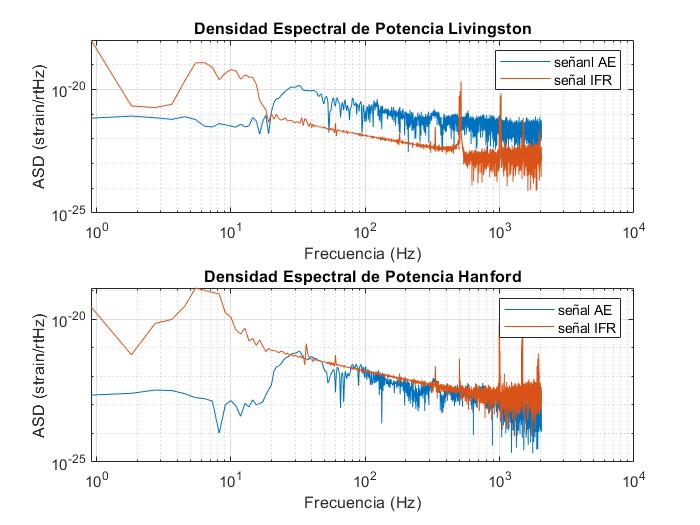}}
\caption{Densidad espectral de potencia del par entrada-salida del autoencoder I.}
\label{fig10}
\end{figure}

\section{Conclusiones}

La aplicación del \textit{autoencoder} en el procesamiento de las señales provenientes de los detectores Hanford y Livingston en torno al evento GW150914 ha demostrado ser una estrategia efectiva para reducir el ruido presente en la señal detectada. Entre las distintas arquitecturas evaluadas, la primera arquitectura del \textit{autoencoder} se destacó por su capacidad para filtrar el ruido de manera óptima.

La disminución notable en la amplitud de ciertas frecuencias significativas por encima de los 1000 Hz, así como la reducción general en las frecuencias bajas, evidencian la capacidad del \textit{autoencoder} para mejorar la calidad de las señales procesadas. 

Con respecto al proceso de filtrado de la señal, no se requirió contar con un conocimiento detallado acerca de la naturaleza de dicha señal ni de sus propiedades espectrales. Esto resulta ventajoso al momento de implementar el algoritmo, en contraste con enfoques previos mencionados en [2] y [3]. En dichos enfoques, el procesamiento depende de la información específica de los datos (asumiendo ciertas características en la distribución espectral) y de un conocimiento exhaustivo sobre el instrumental involucrado. En cambio, el enfoque del presente estudio destaca la capacidad para operar sin requerir supuestos rigurosos acerca de la señal o conocimiento detallado del espectro, lo que lo convierte en una herramienta versátil y adaptable en diversos escenarios de análisis de señales y un buen recurso para introducir en entornos educativos.

Los resultados obtenidos sugieren que la implementación de técnicas de reducción de ruido basadas en \textit{autoencoders} puede resultar útiles en el análisis de señales con múltiples fuentes de interferencia. Sin embargo, se considera que el proceso de \textit{denoising} puede ser optimizado contemplando otras arquitecturas de \textit{autoencoders} que profundicen la arquitectura, implementen redes neuronales recursivas (RNNs) para mejorar la relación señal a ruido, o incluyan una estructura del tipo variacional que permita capturar en el espacio latente una descripción más precisa de la distribución de eventos gravitacionales. No obstante, dichas consideraciones se contemplan para trabajos a futuro.

\vspace{12pt}

\end{document}